\documentclass[sn-mathphys,Numbered]{sn-jnl}

\usepackage{graphicx}
\usepackage{multirow}
\usepackage{amsmath,amssymb,amsfonts}
\usepackage[title]{appendix}
\usepackage{xcolor}
\usepackage{textcomp}
\usepackage{manyfoot}
\usepackage{booktabs}
\usepackage{color}
\usepackage{subcaption}
\usepackage{glossaries}
\usepackage{cleveref}

\newacronym{SSL}{SSL}{Self-Supervised Learning}
\newacronym{FSL}{FSL}{Few-Shot Learning}

\begin{document}
\title{Self-Supervised and Few-Shot Learning for Robust Bioaerosol Monitoring}

\author[1]{\fnm{Adrian} \sur{Willi}}
\author[1]{\fnm{Pascal} \sur{Baumann}}
\author[2,3]{\fnm{Sophie} \sur{Erb}}
\author[1,4]{\fnm{Fabian} \sur{Gr\"oger}}
\author[5]{\fnm{Yanick} \sur{Zeder}}
\author*[1]{\fnm{Simone} \sur{Lionetti}}\email{simone.lionetti@hslu.ch}

\affil[1]{
\orgname{Lucerne University of Applied Sciences and Arts},
\orgaddress{\street{Suurstoffi 4}, \city{Rotkreuz}, \postcode{CH-6343}, \state{ZG}, \country{Switzerland}}%
}
\affil[2]{
\orgname{Federal Office of Meteorology and Climatology MeteoSwiss},
\orgaddress{\street{Chemin de l'A\'erologie 1}, \city{Payerne}, \postcode{CH-1530}, \state{VD}, \country{Switzerland}}%
}
\affil[3]{
\orgname{\'Ecole Polytechnique F\'ed\'erale de Lausanne},
\orgaddress{\street{Station 2}, \city{Lausanne}, \postcode{CH-1015}, \state{VD}, \country{Switzerland}}%
}
\affil[4]{
\orgname{University of Basel},
\orgaddress{\street{Hegenheimermattweg 167b}, \city{Allschwil}, \postcode{CH-4123}, \state{BS}, \country{Switzerland}}%
}
\affil[5]{
\orgname{Swisens AG},
\orgaddress{\street{Meierhofstrasse 5a}, \city{Emmen}, \postcode{CH-6032}, \state{LU}, \country{Switzerland}}%
}

\abstract{
Real-time bioaerosol monitoring is improving the quality of life for people affected by allergies,
but it often relies on deep-learning models which pose challenges for widespread adoption.
These models are typically trained in a supervised fashion
and require considerable effort to produce large amounts of annotated data,
an effort that must be repeated for new particles, geographical regions, or measurement systems.
In this work, we show that self-supervised learning and few-shot learning can be combined
to classify holographic images of bioaerosol particles
using a large collection of unlabelled data and only a few examples for each particle type.
We first demonstrate that self-supervision
on pictures of unidentified particles from ambient air measurements 
enhances identification even when labelled data is abundant.
Most importantly, it greatly improves few-shot classification
when only a handful of labelled images are available.
Our findings suggest that real-time bioaerosol monitoring workflows can be substantially optimized,
and the effort required to adapt models for different situations considerably reduced.
}

\keywords{Pollen, holographic imaging, self-supervised learning, few-shot learning}

\maketitle

\section{Introduction}

Real-time aerosol monitoring provides key information for public health, particularly for managing allergies and asthma \cite{buters2022automatic}.
Pollen is the most common cause of respiratory allergies in European countries, making pollen monitoring key to diagnosing, managing, and treating symptoms.
Although Hirst-type impactors \cite{hirst1952automatic} have been providing reliable airborne pollen data for decades, they have limitations such as low sampling rates (10 l/min), processing delays, and labor-intensive operations.

Recent advances in laser and artificial intelligence technologies address some of these limitations and have lead to the development of new automatic bioaerosol monitoring instruments \cite{huffman2020real}.
The SwisensPoleno is one such new system based on airflow cytometry \cite{sauvageat2020real}.
It measures particles through holographic imaging, laser-induced fluorescence, light scattering and polarisation.
These measurements are then used to classify airborne particles using deep-learning models.
To date, this has been carried out in a setting which requires substantial amounts of labelled data.
The annotation of this data needs to be performed by experts and is labor-intensive, and thus often constitutes a bottleneck.
Another limitation of this approach is that models developed for a certain geographical region
can perform poorly in other regions where atmosphere composition is different.
To classify new types of bioaerosol not included in the training data it is typically necessary to re-train models.
Current research, including European efforts like the EUMETNET AutoPollen Programme \cite{autopollen2018} and the SYLVA Horizon Europe project \cite{sylva2023}, are developing models and networks at the European level to address this issue.

In this paper, we focus on the identification of ambient particles based on holographic images acquired by the SwisensPoleno.
Starting with a general-purpose deep-learning model trained on ImageNet with supervision,
we refine it on 20 million unlabelled images of airborne particles
with a \acrlong{SSL} method called SimCLR.
Using pollen as a case study, we demonstrate that, when a small number of labelled examples are available, the combination of self-supervised training with \acrlong{FSL} is greatly beneficial for particle identification.
We also show that model refinement with self-supervision leads to enhanced robustness to variations in data acquisition settings.

\section{Methods}

This study leverages deep learning to process automatically captured holographic images of ambient particles.
Generalized and robust representations of unlabelled images are constructed using \gls{SSL}.
Subsequently, these representations are combined with \gls{FSL}
to improve classification performance using only a minimal set of labelled samples.

\subsection{Data}

The SwisensPoleno instrument \cite{sauvageat2020real} takes in-flight holographic images of particles as they flow through the measurement system.
Passing particles with sizes between 5 and 200 \textmu m trigger two cameras, which are perpendicular to each other and to the particle flow.
Their raw holographic images are reconstructed in focus, centered,
and cropped to 200$\times$200 pixels grey-scale pictures where one pixel corresponds to 0.595 \textmu m.

In this study, we use both unlabelled and labelled data.
The unlabelled data consists of about 20 million holographic images of ambient airborne particles measured by a SwisensPoleno located on the rooftop of the MeteoSwiss building in Payerne, Switzerland, throughout one year.
The labelled data were obtained from specific measurements during which pollen collected directly from known plants were aerosolized.
These data were then manually filtered and cleaned as described in previous work \cite{erb2023real}. Eleven different plant taxa were considered (\textit{Alnus glutinosa, Carpinus betulus, Cupressus sempervirens, Cynosurus cristatus, Fagus sylvatica, Fraxinus excelsior, Picea abies, Populus sp., Quercus robur, Taxus baccata} and \textit{Ulmus glabra}).
The labelled data were obtained from two SwisensPoleno systems denoted P4 and P5.
Although the large unlabelled collection comes from P5, there is no overlap between the labelled and unlabelled data.

\subsection{Machine Learning}

\begin{figure}
\centering
\includegraphics[width=0.7\linewidth]{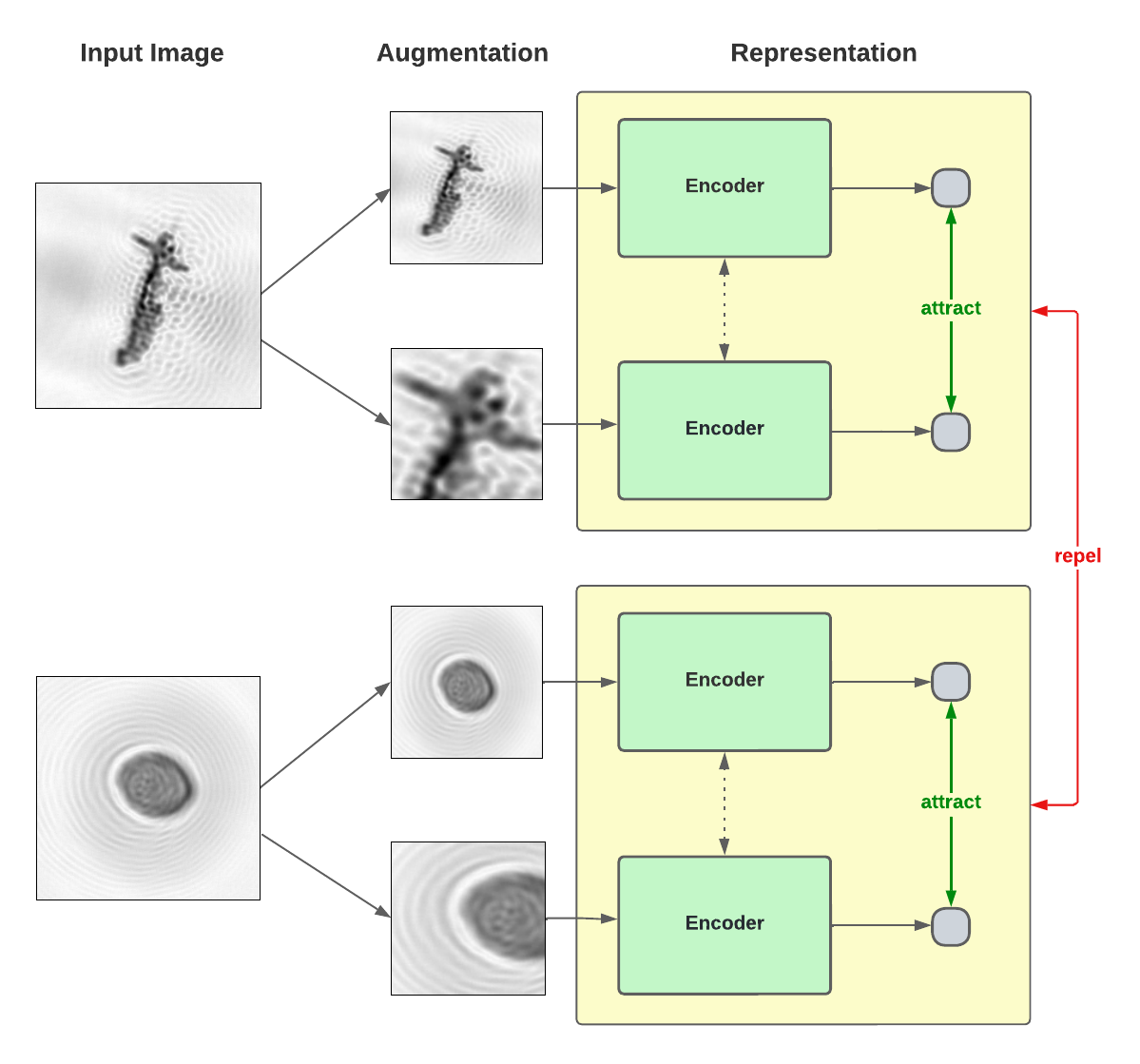}
\caption{
Simplified illustration of the SimCLR framework using holographic images of airborne particles from SwisensPoleno. Every input image in a batch is augmented, resulting in different views of the same particle. These views are then all passed through the same deep neural network encoder to obtain vector representations. Finally, representations from the same input image are attracted while the ones from different input images are repelled. This produces semantic representations without the need for labelled data.
}\label{fig:simclr-data-augmentation}
\end{figure}

\Gls{SSL} offers a way to build models without labelled data.
In particular SimCLR \cite{chen2020simple}, the method used in this work, belongs to the family of contrastive \gls{SSL} approaches.
These methods learn vector representations of images by generating multiple augmented versions for each original image through augmentations like rotation and clipping, and using a deep neural network encoder to map the augmented images into vectors.
The encoder is trained by pulling together these vectors if they come from the same original image and repelling them otherwise, as illustrated in \cref{fig:simclr-data-augmentation}.
We apply SimCLR to SwisensPoleno holographic images and initialize it with a general-purpose image encoder pre-trained on ImageNet in a supervised way, as this performs better compared to random initialization \cite{azizi2023robust}.

To classify new data with a limited number of labelled images, we combine \gls{SSL} with \gls{FSL}.
First, we test a linear classifier, where we combine image representations with one-vs-all logistic regression.
Second, we learn a prototype in representation space for each taxon
by minimizing cross-entropy after assigning samples to prototypes with cosine similarity.
The latter approach was proven to perform consistently better than the linear classifier
and on par with more sophisticated techniques across many \gls{FSL} scenarios \cite{chen2019closer}.

We compare the model trained on ImageNet
to the one that has been specialized for holographic aerosol images with SimCLR,
always using the same EfficientNet-B0 architecture \cite{tan2019efficientnet}.
We measure performance using balanced accuracy as our dataset does not reflect atmospheric composition.

\section{Results}

\begin{table}
    \centering
    \caption{
    Balanced accuracy for the classification of 11 pollen taxa from the features of the ImageNet+SimCLR model and ImageNet, using a linear classifier (one-vs-all logistic regression) trained on a labelled P5 training dataset and evaluated on the labelled P5 and P4 test datasets.
    }
    \label{tab:frozen-eval-lr}
    \begin{tabular*}{\linewidth}{@{\extracolsep{\fill}}lcc} 
    \toprule
    & \multicolumn{2}{c}{\textbf{Balanced Accuracy Score}}\\
    \textbf{Method} & \textbf{P5 test dataset} & \textbf{P4 test dataset}\\  
    \midrule 
    ImageNet & 89.9 \% & 75.5 \% \\
    ImageNet+SimCLR & 90.4 \% & 79.0 \% \\
    \botrule
    \end{tabular*}
\end{table}

We first consider the case where the model is trained on all available labelled data.
We use 400 to 6'600 pollen grains per taxon for training and evaluate previously learned representations with one-vs-all logistic regression.
Results are presented in \cref{tab:frozen-eval-lr}.
When evaluating the model trained on data from the same instrument P5, we observe that specializing the supervised ImageNet features on SwisensPoleno data with SimCLR marginally improves the balanced accuracy (+0.5\%) compared to ImageNet alone.
However, the gain is larger (+3.5\%) when evaluation is carried out on data from the other instrument P4.
In these conditions, the ImageNet performance drops to 75.5\% while that of ImageNet+SimCLR only drops to 79\%.

\begin{figure}
\centering
\includegraphics[width=1\linewidth]{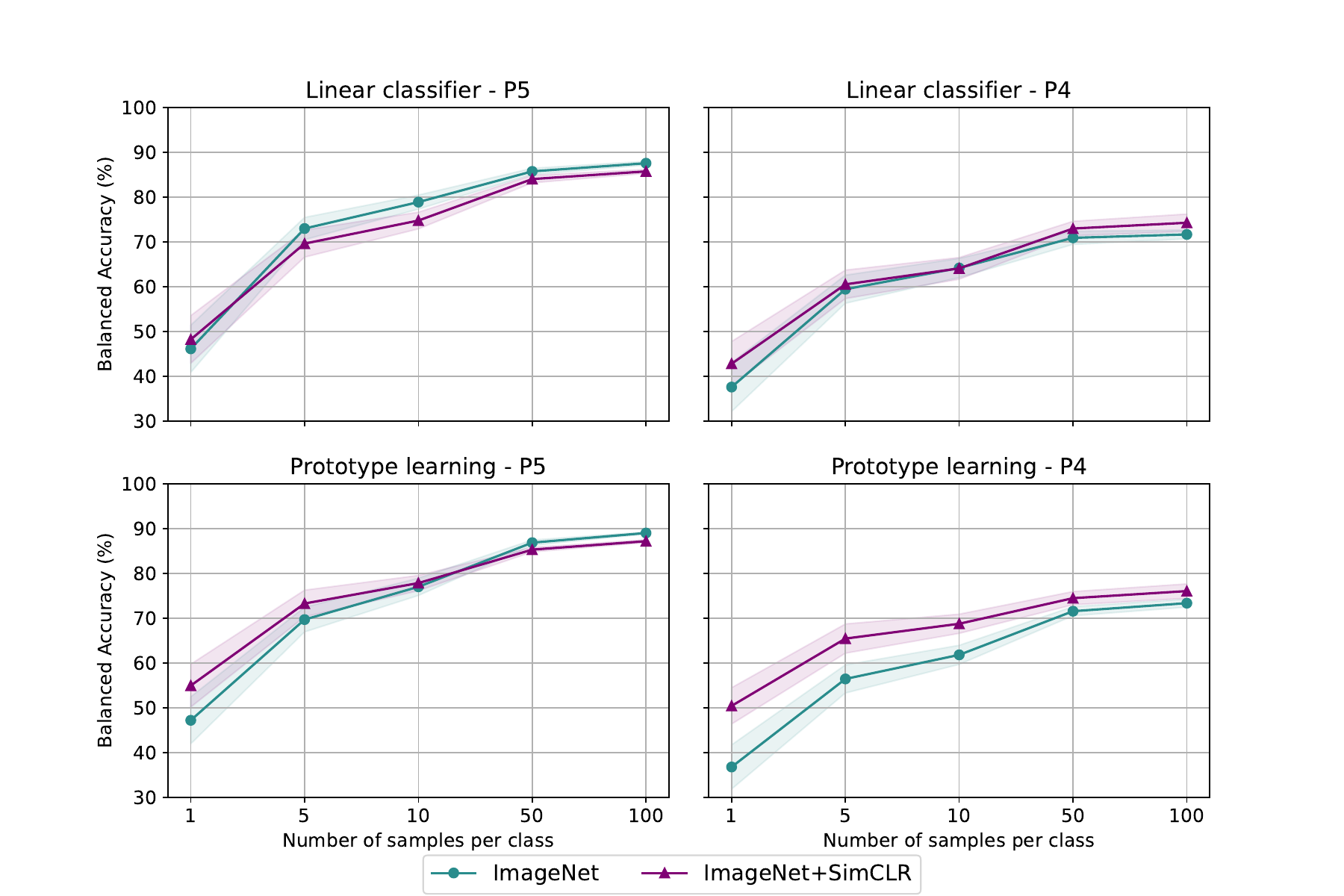}
\caption{
Balanced accuracy for pollen classification as a function of the number of available labelled images per taxon, evaluated on data from the same SwisensPoleno (P5) used for \gls{SSL} (left), and on data from another SwisensPoleno (P4) (right). Results are shown for the linear classifier (top) and prototype learning (bottom) on top of features from ImageNet (teal) and ImageNet+SimCLR (purple). Colored bands represent the uncertainty due to the random choice of the labelled images.
}
\label{fig:few-shot-plots}
\end{figure}

Next, we address situations where training data are scarce.
\Cref{fig:few-shot-plots} shows the performance of different methods as a function of the number of labelled images available per taxon.
When evaluation is performed on test data from the same measurement system as the training data,
i.e.\ from P5 (left column),
ImageNet+SimCLR outperforms ImageNet with only one image per taxon.
As the amount of labelled data increases, ImageNet gains an edge over ImageNet+SimCLR,
and differences among models become smaller.
If less than 10 images per taxon are available,
the best results are obtained by combining ImageNet+SimCLR with prototype learning,
while the ImageNet model does not profit from this \gls{FSL} technique.
When the evaluation is performed on test data from a different measurement system
compared to the training data, i.e.\ P4 (right column),
ImageNet+SimCLR outperforms ImageNet throughout.
Moreover, the synergy of ImageNet+SimCLR and prototype learning yields significantly better results
compared to the other three alternatives.
When a single example is available, this combination achieves 50\% balanced accuracy
while all other methods are around 40\%.
With only 5 or 10 labelled examples, one can already expect a balanced accuracy around 65\% and 70\%
respectively.

In general, we observe significant performance drops
when evaluation is performed in a different setting compared to training,
even if holographic images exhibit very little variations among measurement systems.
These drops are markedly reduced in the case of ImageNet+SimCLR with prototype learning,
highlighting the robustness of distance-based classification with self-supervised features.
By contrast, purely supervised models are more prone to take shortcuts for classification,
which might explain why they generalise poorly when conditions change \cite{robinson2021can}.

\section{Discussion and conclusion}

ImageNet+SimCLR offers more robust performance when models are evaluated under different conditions compared to training.
This is already an advantage for pollen monitoring networks.
However, the combination of \gls{SSL} with \gls{FSL} can produce flexible identification models,
and therefore has an even greater appeal for networks that cover a large geographical region.
Although \gls{SSL} requires additional training compared to off-the-shelf models,
this only needs to be done once and taxa can be specified a posteriori.
As long as the initial \gls{SSL} training dataset is broad enough, the model learns features to distinguish arbitrary particles, and good performance is expected with \gls{FSL} for any new taxon.
The significant potential benefit for the development of new pollen identification models is that the effort of labelling pollen data may be reduced to a handful of particles, if performance remains within acceptable limits.
This opens the door to faster development of models across climatic regions.

In this study, we showed that self-supervision can effectively leverage large volumes of unlabelled data to improve the robustness of deep-learning models for pollen grain identification, not only against geographical variation but also against changes in data collection conditions.
The approach makes data preparation more feasible and produces flexible models, since other bioaerosol types can be classified with \gls{FSL} as long as the \gls{SSL} training dataset is sufficiently diverse.
Results could be improved further by training the model on data which covers different collection conditions, including several instruments and different geographical regions.
Similarly, incorporating data from other sensors could enhance identification accuracy.
Future work should include testing other \gls{SSL} methods, especially non-contrastive ones \cite{grill2020bootstrap}, and merging current research efforts to develop a foundation model handling a large set of bioaerosol.

\backmatter

\section*{Declarations}

\subsection*{Funding}
The data used in this work was provided by the Federal Office of Meteorology and Climatology MeteoSwiss.
PB, FG, and SL gratefully acknowledge the support of the Swiss Innovation Agency Innosuisse
through preliminary study voucher 59965.1 INNO-ICT.
SE was funded by the Swiss National Science Foundation (IZCOZ0\textunderscore198117).

\subsection*{Competing interests}
YZ is an employee of Swisens AG, and
SL discloses support from Swisens AG covering the supervision for AW’s master thesis
as foreseen by the regulations of the Lucerne University of Applied Sciences and Arts.
All investigations were carried out in compliance with good scientific practices
and the declared relationships have no effect on the results presented.

\subsection*{Availability of data and materials}
The data presented in this paper were kindly provided by the Federal Office of Meteorology and Climatology MeteoSwiss, and cannot be shared publicly.

\subsection*{Code availability}
The code used in this work is experimental and tailored to the SwisensPoleno.
Considering that it cannot be used without the specific data to reproduce results nor for applications with other devices, there is little benefit making it open-source.

\subsection*{Authors' contributions}
SL, YZ, and FG planned the work.
SE provided the data.
AW, PB, FG, and SL contributed to the analysis.
AW, SE, SL, and YZ wrote the manuscript, and FG and PB revised it.

\bibliography{bibliography}

\end{document}